\documentclass[sigconf]{aamas} 
\usepackage{babel}

\usepackage{balance} 

\makeatletter
\gdef\@copyrightpermission{
  \begin{minipage}{0.2\columnwidth}
   \href{https://creativecommons.org/licenses/by/4.0/}{\includegraphics[width=0.90\textwidth]{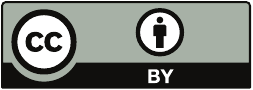}}
  \end{minipage}\hfill
  \begin{minipage}{0.8\columnwidth}
   \href{https://creativecommons.org/licenses/by/4.0/}{This work is licensed under a Creative Commons Attribution International 4.0 License.}
  \end{minipage}
  \vspace{5pt}
}
\makeatother

\setcopyright{ifaamas}
\acmConference[AAMAS '25]{Proc.\@ of the 24th International Conference
on Autonomous Agents and Multiagent Systems (AAMAS 2025)}{May 19 -- 23, 2025}
{Detroit, Michigan, USA}{Y.~Vorobeychik, S.~Das, A.~Nowé  (eds.)}
\copyrightyear{2025}
\acmYear{2025}
\acmDOI{}
\acmPrice{}
\acmISBN{}

\acmSubmissionID{499}

\title[The Many Challenges of Human-Like AI]{The Many Challenges of Human-Like Agents \\ in Virtual Game Environments}

\author{Maciej {\'S}wiechowski} 
\affiliation{
  \institution{QED Software}
  \city{Warsaw}
  \country{Poland}}
\email{maciej.swiechowski@qed.pl}

\author{Dominik {\'S}l\k{e}zak} 
\affiliation{
  \institution{University of Warsaw}
  \city{Warsaw}
  \country{Poland}}
\email{slezak@mimuw.edu.pl}

\begin{abstract}
Human-like agents are an increasingly important topic in games and beyond. Believable non-player characters enhance the gaming experience by improving immersion and providing entertainment. They also offer players the opportunity to engage with AI entities that can function as opponents, teachers, or cooperating partners. Additionally, in games where bots are prohibited -- and even more so in non-game environments -- there is a need for methods capable of identifying whether digital interactions occur with bots or humans. This leads to two fundamental research questions: (1) how to model and implement human-like AI, and (2) how to measure its degree of human likeness.

This article offers two contributions. The first one is a survey of the most significant challenges in implementing human-like AI in games (or any virtual environment featuring simulated agents, although this article specifically focuses on games). Thirteen such challenges, both conceptual and technical, are discussed in detail.

The second is an empirical study performed in a tactical video game that addresses the research question: ``Is it possible to distinguish human players from bots (AI agents) based on empirical data?'' A machine-learning approach using a custom deep recurrent convolutional neural network is presented. 

We hypothesize that the more challenging it is to create human-like AI for a given game, the easier it becomes to develop a method for distinguishing humans from AI-driven players.
\end{abstract}

\keywords{Intelligent Agents; Human-Like AI; Believable Agents; ML Model for Turing Test}

\newcommand{\BibTeX}{\rm B\kern-.05em{\sc i\kern-.025em b}\kern-.08em\TeX}

\begin{document}


\pagestyle{fancy}
\fancyhead{}


Copyright © 2025 by International Foundation for Autonomous Agents and Multiagent Systems (IFAAMAS). Permission to make digital or hard copies of portions of this work for personal or classroom use is granted without fee provided that copies are not made or distributed for profit or commercial advantage and that copies bear this notice and the full citation on the first page. Copyright for components of this work owned by others than IFAAMAS must be honored. Abstracting with credit is permitted. To copy otherwise, to republish, to post on servers or to redistribute to lists, requires prior specific permission and/or a fee.

BibTeX:

\begin{verbatim}
@inproceedings{humanLike2025,
  title={{The Many Challenges of Human-Like Agents in Virtual Game Environments}},
  author={{\'S}wiechowski, Maciej and {\'S}l{\k{e}}zak, Dominik},
  booktitle={Proc. of the 24th International Conference on Autonomous Agents and Multiagent Systems (AAMAS'25)},
  year={2025},
  doi={10.5555/3709347.3743837},
  isbn = {9798400714269},
  publisher={IFAAMAS},
  pages={1996--2005},
  series = {AAMAS '25}
}
\end{verbatim}
\maketitle 


\begin{acks}
This research was co-funded by the Smart Growth Operational Programme 2014-2020, financed by the European Regional Development Fund under a GameINN project POIR.01.02.00-00-0207/20, operated by The National Centre for Research and Development.
\end{acks}

\section{Introduction}

Games have been an integral part of human civilization since ancient times~\cite{kurke1999ancient}. In the essay ``Homo Ludens'', the author examines games as a fundamental condition for the evolution of culture~\cite{ehrmann1968homo}. Besides providing entertainment, some games help in training the mind, enhancing eye-brain coordination, and improving reflexes. 

Since the advent of Artificial Intelligence (AI), games have served as a testbed for its development. Initially, AI research focused primarily on \emph{chess}~\cite{mccarthy1990chess} and \emph{checkers}~\cite{samuel1959some}. However, with recent advancements that have sparked debates about the human-likeness of AI, video games have become increasingly attractive as research environments. Unlike abstract combinatorial games such as \emph{chess}, \emph{checkers}, or \emph{Go}~\cite{gelly:go}, video games typically offer a simplified model of the real world populated with numerous non-player characters (NPCs). They were named ``\textit{human-level AI's killer application}''~\cite{Laird2001}.

The aim of this article is twofold and is divided into two parts.
The first part, contained in Section~\ref{sec:challenges}, provides a comprehensive and accessible review of the challenges related to creating human-like AI (characters, bots, players) for games. We discuss 13 challenges, drawing from both the literature and our years of professional experience in implementing artificial intelligence for video games. These challenges can be generalized to autonomous robots and any virtual environments with intelligent agents. We believe that the comments and insights provided mostly in this part can help researchers design AI that acts more human-like by addressing the issues related to creation and evaluation.
In our search for relevant papers, we queried popular bibliographic databases using terms from the following template: $\langle$ adjective $\rangle$ $\langle$ noun $\rangle$. 
The set of adjectives included ``human-like'', ``human-level'', ``believable'', and the set of nouns contained: ``agent(s)'', ``player(s)'', ``bot(s)'',  ``character(s)'', ``behavior'', ``AI'', and ``Artificial Intelligence''. The complete set of terms was a Cartesian product of the sets of adjectives and nouns. Additionally, we included other selected articles such as~\cite{silver2016mastering} that are seminal to the topic. Among all the papers matching the query, we read their abstracts and introductions to verify if their contents were related to creating or evaluating human-like AI in virtual environments. After this step, $54$ papers remained. We have analyzed them and distilled the most pertinent and common challenges.

Parallel to creating human-like autonomous agents, we also discuss the issue of assessing their human-likeness. These topics are closely intertwined. It is challenging to implement AI techniques without precisely setting the evaluation criteria—what constitutes human-likeness. 
In 1950, Alan Turing posed the question ``Can Machines Think?''~\cite{turing}. He introduced the concept of the \emph{Imitation Game}, which laid the foundations for what would later be known as the \emph{Turing Test}~\cite{turing-analysis}. In the classic variant of the test, an interrogator (judge) interacts with two players, \emph{A} and \emph{B}, using a natural language chat interface in such a way that the players cannot be seen—only their responses can. The goal of the interrogator is to determine whether each participant is a human or a computer. The \emph{Turing Test} has been a foundational concept in AI and has sparked a debate about whether machines can display human-like abilities, intelligence, and consciousness. This topic has been pursued by many prominent researchers, such as Lofti Zadeh~\cite{zadeh2008toward}. 
Although there have been attempts to formalize believability~\cite{bogdanovych2015formalising}, the most common approach is to propose a \emph{Turing Test} analogy for video games~\cite{swiechowski2020game}. The first research framework to do so was the \emph{2K BotPrize}~\cite{hingston2009turing}, proposed in 2008 by Philip Hingston, based on the multi-player shooter game \emph{Unreal Tournament 2004}. Since then, researchers have adopted the idea of the \emph{Turing Test} for more games, such as \emph{Street Fighter}~\cite{arzate2018hrlb} and \emph{Ms. Pacman}~\cite{Miranda2017PacManOP}. 

In the second part, we present a study concerning the automatic construction of a method capable of distinguishing human players from bots solely by learning from data. In Section~\ref{sec:environment}, the environment in which the experiments were conducted is presented—a tactical war video game with relatively high action-space complexity. Section~\ref{sec:method} focuses on the machine learning algorithm proposed for this task. The training methodology combines recurrent and convolutional neural networks and utilizes multi-modal (numeric and spatial) data. The proposed solution achieves an F1-Score of $0.92$, which is a significant improvement over the previous solution based on XGBoost with only numeric features, which had an F1-score of $0.58$. The last section is devoted to conclusions.

\section{Discussion of the Challenges}
\label{sec:challenges}

\noindent \textbf{Humans are Diverse}. The first challenge we wish to highlight is the ambiguity and imprecision inherent in defining the goal of human-like AI. Following \cite{arzate2018hrlb}, human-like AI is described as ``behaving in a manner that makes it indistinguishable from human players''. The authors of~\cite{tence2010challenge} define it as ``giving the feeling of being controlled by a (human) player''. Considering the vast diversity among humans, what exactly does it mean to play like a human player? The playing styles can vary significantly. An elderly individual often plays differently compared to a younger person or a child who just started playing video games. Similarly, a professional player's approach will differ from that of an amateur. Even among seasoned players, there is a lot of variety. The authors in~\cite{arinbjarnar2012actor} identify seven distinct player types: \emph{power gamer}, \emph{butt-kicker}, \emph{tactician}, \emph{specialist}, \emph{method actor}, \emph{storyteller}, and \emph{casual gamer}. Another example is the Bartle taxonomy, outlined in~\cite{zuchowska2021bartle}, that categorizes players into four roles: \emph{killer}, \emph{socializer}, \emph{achiever}, and \emph{explorer}. Each category adjusts their gameplay in a unique manner to align with their personal objectives and maximize enjoyment.

\textit{Unreal Tournament 2004} serves as a popular choice among researchers for assessing human-like behavior in gameplay. According to \cite{gamez2012neurally}, one of the significant challenges for human judges is the wide range of human behaviors, which underscores the complexity due to player diversity. The authors investigated the topic further and commented that skill level is the most crucial factor distinguishing different playing patterns. In their study, less skilled players were identified as outliers in a statistical analysis of gameplay when matched with experts.

Additionally, implementing AI bots to mimic human behavior can be problematic if the game is still under development, because it might be unclear how human players will play the particular game.  This becomes particularly challenging for novel games, i.e., not based on standard repeatable formats.\\

\noindent \textbf{The Complexity and Expressiveness of Action Space}. In short, complex high-dimensional action spaces pose a great challenge in implementing human-like AI, whereas simple, constrained ones complicate the evaluation of whether an AI is truly human-like. Let us elaborate on this.

In complex environments, human-like behavior necessitates the simulation of thinking processes. These usually involve various forms of reasoning and strategic or tactical planning that anticipate multiple steps ahead. Humans often develop behavior patterns based on their experiences and can rely on their intuition. In contrast, AI-driven characters depend on computational techniques such as tree search (e.g., alpha-beta or Monte Carlo Tree Search (MCTS)~\cite{mcts}), planning (such as Hierarchical Task Networks (HTN)~\cite{nejati2006learning}), or machine learning. The complexity of the environment significantly increases the computational resources required to create even a moderately proficient AI player compared to human players. This is clearly demonstrated by historical attempts to challenge top human players in various games, such as \emph{Chinook}~\cite{schaeffer2007checkers} (in checkers), \emph{Deep Blue}~\cite{campbell2002deep} (in chess), \emph{AlphaGo}~\cite{silver2017mastering} (in Go), \emph{AlphaStar}~\cite{alphastar} (in Starcraft II), and \emph{OpenAI Five}~\cite{dotaOpenAI} (in Dota 2). These projects employed substantial computational resources. For instance, Deep Blue was ranked 259th on the top500 list of supercomputers, while OpenAI Five utilized 172,800 CPUs and 1536 GPUs simultaneously~\cite{dotaOpenAI} at a peak. Although creating a potent computer agent is not equivalent to developing human-like AI, these research projects began their efforts when the top programs were significantly inferior to the skills of amateur human players. In complex games, achieving a skill level comparable to an average human player also demands considerable computing power and/or extensive simulation times. The complexity originates from several aspects: the number of options available in each state (termed the branching factor), the length of the game (which necessitates longer simulations, thereby reducing the number that can be conducted within a given time), the diversity of game states (state-space complexity), and the total number of leaf nodes in a complete game tree (game-tree complexity). Chess, being the second least complex game in this context (after checkers), possesses an average branching factor of $35$, a state-space complexity of $10^{44}$, and a game-tree complexity of $10^{123}$. To provide a sense of scale, it is estimated that there are $10^{80}$ atoms in the observable universe. The complexity escalates further in real-time games with continuous movement making the action space virtually infinite (as bots can go in any directions). In the study concerning believable navigation~\cite{karpov2012believable}, the authors found that AI players can get stuck on level geometry or fail to appear human when following the navigation graph. The authors of~\cite{arzate2018hrlb} emphasize the necessity of exploring high-dimensional action spaces as one of the two main challenges in creating human-like AI. 

The five aforementioned AI players were the outcomes of large-scale research projects focused on competing against top human players in dedicated matches. However, AI players integrated into commercial video games must operate within the constraints of a single consumer-level hardware system shared with other game functionalities such as rendering~\cite{rabin2013game}. This restriction makes it particularly challenging to develop AI players equalling humans in skills without giving them an unfair advantage.\\

Now, let us focus on simple abstract game environments characterized by low complexities in both state space and action space. \emph{Tic-Tac-Toe} serves as a good example. While it is straightforward to create competent or even exceptionally strong AI players for such games, differentiating whether a player is human-like proves challenging.  If there are only a handful of actions and the game follow strict rules, e.g., players take turns and each turn the active player chooses one of the few available actions, then there are insufficient premises to distinguish humans from bots. This issue is even more pronounced in \emph{Rock-Paper-Scissors}, where each action is equally viable, making even random choices a legitimate strategy. In the study~\cite{Miranda2017PacManOP}, the authors highlighted the difficulties in telling whether a player is human or a bot within the game \emph{Ms. Pac-Man}, which is considerably more complex than \emph{Tic-Tac-Toe}.

The underlying point is that a game or virtual environment must possess a sufficient level of expressiveness to enable feasible assessments of human-like behavior. The more open-ended an environment is, the easier it becomes to determine whether the behavior of agents within it is believable.\\

\noindent \textbf{The Challenges of Scale.} As mentioned in the previous paragraph, creating human-like AI is extremely computationally demanding. Note that all examples of AI agents that achieved high efficacy were in 1-vs-1 player settings. In these cases, the AI, utilizing powerful hardware, controlled one player (or, at most, two during training via self-play).

Now, imagine a virtual city populated by a million NPC inhabitants, each individually controlled. These NPCs must take actions based on various circumstances such as their goals and both internal and external states. One of the models for simulating many agents is called Belief–Desire–Intention (BDI)~\cite{georgeff1999belief}. Scaling high-efficacy AI implementations to environments with many diverse agents presents a challenge that is multiple orders of magnitude greater~\cite{Bailey2012}. Essentially, it requires multiplying the resources used to create a high-fidelity solution by the number of simulated agents. As of the time of this article, it is infeasible to apply \emph{state-of-the-art} game-playing models to massively multiplayer games. Currently, dedicated, low-fidelity approaches known as ``Crowd AI'' are used in the industry. Nevertheless, there have been research attempts to tackle the problem of simulating a large number of believable bots, as discussed by \cite{rankin2010scalable}. This approach utilizes Utility AI, integrates the Observe-Orient-Decide-Act (OODA) decision cycle, and employs temporary roles that agents may assume.\\

\noindent \textbf{Avoiding Superhuman Behavior}. Rational players must act towards their goals and posses a certain level of skill in order to be believable. In efforts to develop AI agents with sufficient competence, there is a risk that these agents may display superhuman abilities in certain aspects of a game, compromising their believability. Both~\cite{cavazza2000computer} and~\cite{livingstone2006turing} highlight that precision in calculations, such as pixel-perfect aiming in shooter games, is a characteristic strongly associated with bots in scenarios analogous to the \emph{Turing Test} for video games. Therefore, it is another feature that makes it easier to assess human-like behavior but more difficult to implement it. It is often the case, that the removal of such precise calculations from bots make them significantly weaker, to the point that they are not believable due to different reasons, i.e., their incapacity to perform competent actions.\\

\noindent \textbf{Idle and Non-Relevant Actions}. Even when a game has a clearly defined objective, human players often engage in actions that are irrelevant from the perspective of this goal. Such behaviors are commonly described in the literature as roaming, idle, ``for fun'', or for ``own amusement'' actions. Observing navigation in an open virtual world provides a particularly interesting context to study these patterns~\cite{milani2023navigates}. For example, players exploring a city in an RPG might pause to admire specific 3D models within the game environment. The following example, based on the author's experience in a \emph{Diablo} game developed by \emph{Blizzard Entertainment}, illustrates this point. Imagine a scenario where a player controls a character equipped with an artifact that leaves a trail of ice on the ground. Players might intentionally run in patterns to create specific shapes with this mechanic. This emergent behavior, stemming from human creativity and playfulness, is entirely disconnected from the game's mechanics or objectives. Consequently, it presents significant challenges for implementing such behaviors in AI-controlled bots. Standard algorithms designed to enhance AI player proficiency, which is typically the primary focus, would likely treat these actions as noise -- something irrelevant. The study mentioned in~\cite{10.5555/3060832.3060973} proposes a modification to the Monte Carlo Tree Search (MCTS) algorithm that biases action selection towards patterns observed in human gameplay to make it more human-like. Given that enough data is available from human players, this adaptation appears to be a promising strategy for developing more human-like AI behaviors.\\

\noindent \textbf{Biological Constraints}. Equipping AI players with realistic biological constraints extends well beyond standard practices in the field~\cite{rabin2013game, millington2019ai}. These constraints include:

\begin{itemize}
\item \emph{Visual Perception}. Typically, AI-controlled players are fed information directly about game objects and other characters, without any simulation of actual perception. For example, one of the distinguishing factors between bots and humans in UT2004 is that bots often collect items not looking at them~\cite{Schrum2012}. The article~\cite{laird2002} discusses the challenge of extracting spatial information about the physical environment -- such as walls and doors -- from the game's internal data structures, which are just sets of polygons.

As an interesting side note, the authors of~\cite{togelius2012assessing} provided an insightful finding in their paper. The assessment of a bot's human-likeness, as judged by humans, showed significant variation depending on the observation perspective. The authors concluded that a more accurate assessment occurred when bots were observed from a third-person perspective, as opposed to a first-person perspective.

\item \emph{Sound Perception}. Simulating realistic sound perception is as challenging as visual perception. The detection of sounds by a bot should not be binary, it should involve some level of fuzziness and imperfection, similar to the senses of living organisms. For instance, the bot should be able to express uncertainty by saying, ``I didn't hear you.''

\item \emph{Memory}. In the article ``Do Non-Player Characters Dream of Electric Sheep?''~\cite{johansson2013non}, the author emphasizes memory functions as a crucial component for creating believable NPCs. Designing a memory system that encompasses prioritization of information, realistic forgetting, and loss of detail while retaining key facts presents significant challenges. These aspects must be managed without undermining the goal-oriented performance of AI players, a recurring theme in this section.

\item \emph{Cognitive Load and Ability to Multitask}. In real-time video games, especially strategic ones, AI players often display less strategic reasoning than humans. However, they compensate by being able to oversee the entire map and control numerous units at once, which would be overwhelming and physically impossible for human players. Simulating realistic constraints related to attention focus and multitasking poses multiple challenges: determining appropriate game-specific limits, deciding which aspects of the game the AI should focus on, and the fact that imposing such constraints might diminish the AI’s effectiveness. Creating competent bots that operate on consumer-level hardware remains a significant challenge.

\item \emph{Reaction Time}. Reaction time serves as another biological constraint. In video games, AI players frequently benefit from seemingly unlimited reaction times, a key feature distinguishing bots from human players~\cite{Mandziuk2012}. There have been propositions to limit the reaction time and action rate of AI players, as seen in~\cite{alphastar}.

\item \emph{Other Constraints}. The article titled ``Video Game Agents with Human-like Behavior using the Deep Q-Network and Biological Constraints''~\cite{morita2023video} introduces additional constraints such as ``confusion'' (experienced when suddenly surrounded by many enemies), ``fluctuation'' (errors in operation), ``delay'' (akin to reaction time), and ``tiredness''.
\end{itemize}

\noindent \textbf{Emotional Element.} The authors of~\cite{elsayed2017affect} wrote:

\begin{quote} ``\textit{Problems of NPCs usually lie in their lack of convincing
social and emotional behaviour raising the need for a robust
affect module within the agent’s architecture. Developing an
integrated architecture would ideally require developing
models for the theory of emotion, social relation, and
behaviour, and combining the theories into an overall model.}''
\end{quote}
In~\cite{lee2012you}, emotions are listed as one of the most critical qualities of believable bots, ranking just after goals. The authors of~\cite{bogdanovych2016makes} identify emotions as one of the seven believability characteristics, alongside personality, self-motivation, change, social relationships, consistency of expression, and illusion of life.

Emotions have been extensively studied in psychological research~\cite{izard2013human}. However, there are currently no established computational models to accurately simulate emotions in video games, making this issue particularly challenging. There are many facets to emotions in games. An emotional response may lead to changes in a player's behavior following specific in-game events. The next example will be from the game \emph{Tactical Troops: Anthracite Shift}, discussed in a study in the next section of this article. In this game, a player wins after eliminating all enemy units. We observed a scenario where an AI-controlled player had two units remaining, and the opposing player had only one. One of the AI's units had a clear shot at the enemy but was blocked by another friendly unit. The AI chose to fire anyway, eliminating both the friendly and the enemy unit, thus securing a win. This behavior, while effective, is rarely observed in games played by humans, who typically avoid harming their own units.

Another aspect of emotional display in gaming is the anger or ``tilt'' effect~\cite{tilt}, which refers to a state of emotional frustration or upset that negatively impacts a player's performance. This phenomenon occurs when players become agitated due to a series of losses, perceived unfairness, or other in-game setbacks, leading to increasingly poor decision-making and potentially aggressive behavior. The term ``tilt'' originates from poker but has become widely used across various competitive gaming genres.\\

\noindent \textbf{Handling Uncertainty}. Uncertainty in games typically involves hidden information (asymmetric information between the players), randomness, or both. This already poses significant challenges for creating agents aimed at achieving effective play because it results in a combinatorial explosion of potential game states. In research related to game AI, this issue is generally addressed using determinization techniques~\cite{cowling2012ensemble} and by replacing perfect information game states with information sets~\cite{cowling2012information}. In the video game industry, the amount of hidden information is often so vast that AI characters are usually given unfair access to it, albeit with some techniques implemented to make this less apparent~\cite{liden2003artificial}. For example, in real-time strategy (RTS) games, most of the map is concealed from the players by the so-called ``fog of war.'' Players can send units to anywhere on the map to scout and reveal the covered areas. AI players, however, are typically given hints about a few potential locations of the human player's base, including the correct one, thereby reducing the amount of scouting required. The same applies to information about the resources and military capabilities the human player possesses.

Uncertainty introduces an additional layer of challenge when designing human-like bots. If bots perform too efficiently, for instance, by employing complex probability estimations, they might be perceived by human players as cheating AI~\cite{laird2002}, even if this is not the case. To counteract this, a solution is to integrate a human-like reasoning process for inferring hidden information~\cite{marques2013towards}. However, this approach is computationally intensive and requires significant trade-offs in terms of playing efficacy.\\

\noindent \textbf{Adaptability.} The article~\cite{livingstone2006turing} hypothesizes that AI in commercial games is often exploitable due to its tendency to follow repetitive patterns, thereby allowing human players to recognize and adapt to these behaviors. Meanwhile, the authors of~\cite{evolving} assert that ``adding some unpredictability can significantly enhance believability,'' and the authors of~\cite{soni2008bots} demonstrate a strong correlation between unpredictability and human-like behavior.

In~\cite{arzate2018hrlb}, the authors identify ``adaptation'' to the opponent's style as one of the two primary challenges in developing human-like AI for the \emph{Street Fighter IV} game. The other challenge they outline is exploring a high-dimensional state-action space. To address this, they implemented real-time reward transformations within their reinforcement learning framework. This strategy involved dynamically altering the reward definitions based on the performance of various playing styles. As a result, they achieved a human-likeness score of $0.64$ on a scale from $0$ to $1$, slightly less than the top score of $0.67$ achieved by a human player, significantly surpassing the baseline bots, which scored between $0.28$ and $0.30$.\\

\noindent \textbf{Making Mistakes.} The analysis of literature related to the creation and evaluation of human-like AI underscores that human players indeed make mistakes. As noted by the authors of~\cite{elsayed2017affect}: \begin{quote} ``\textit{An intelligent agent model should not require producing a
``perfect'' agent, but rather, for better human resemblance and
higher believability, it is more natural to have the flaws and
dysfunctionalities of the human affect phenomena
incorporated into the model.}'' \end{quote}

However, humans learn, and there is significantly less likelihood of repeating the same mistakes. In the \emph{UT2004} competition, bots that repeated the same mistakes consistently were quickly judged as non-human-like~\cite{Schrum2012}.

Implementing both the inclusion of mistakes and learning from them is challenging. Mistakes can emerge unintentionally without being explicitly programmed into bots, but these unintentional errors can be challenging to identify and incorporate with adaptation mechanisms. In addition, they can be of artificial nature such as bots getting stuck in level geometry~\cite{Schrum2012}, which was a telling factor for judges responsible for the video game \emph{Turing Test} in \emph{UT2004}. Deciding which aspects of gameplay should intentionally involve mistakes in a believable manner remains a significant challenge.\\

\noindent \textbf{Training Human-Like AI.} One of the most promising approaches to creating human-like AI agents is training them using machine learning methods (ML). When the objective is to develop the strongest agent possible, training can be conducted using reinforcement learning without human knowledge \cite{silver2017mastering, wang2009creating}. In this setup, agents compete against different versions of themselves, continually refining their skills. However, when aiming for human-likeness or believability, training -- whether supervised or through reinforcement -- using human games becomes a more suitable approach~\cite{arzate2018hrlb}.

Training believable agents from human data presents many of the already mentioned challenges: 
\begin{itemize} 

\item \textit{Humans are diverse}. Training can utilize games played by either a specific group of players or the general population. In the latter case, the resultant behavior will likely be an average. 

\item \textit{High computational demand}. Vast action spaces, especially in reinforcement learning, are known for their sample inefficiency and consequent computational expense \cite{hessel2018rainbow}. 

\item \textit{The effect of scale}. The trained model must be inferred as often as there are intelligent agents in the environment. However, modern GPU-based ML models facilitate batch processing. \end{itemize}

Additional challenges specific to training include: 
\begin{itemize} 

\item \textit{Large volumes of training data are required}. This data is usually not available for games at the time of development. As the game has not yet been released, the only data available typically comes from developers and testers playing it, which is insufficient for large-scale training. 

\item \textit{Human-Like Control Input}. In video game development, AI systems typically process inputs differently from human players, which poses a fundamental challenge in creating believable behaviors. Humans use controllers such as: keyboard, mouse, game-pads (including analog joysticks), steering wheels, etc. Using a controller involves atomic actions, e.g., pressing a key or applying a force in one or more of the controller axes. The observable  in-game behaviors emerge from sequences of these inputs, adding complexity to the input space. In contrast, AI in games is usually programmed to perform high-level behaviors like "move to point X", "attack enemy Y", or "flee", which operate on a more abstract level than direct input actions. 
\end{itemize}

\noindent \textbf{Simulating Social Norms.} In the literature related to human-like AI agents, the simulation of social norms is frequently highlighted as an important factor for enhancing believability~\cite{johansson2013non, Weiss2012, warpefelt2013analyzing}. In~\cite{johansson2013non}, the term ``socially believable'' is used to describe agents that effectively cooperate and coordinate within a group. The authors in~\cite{Weiss2012} enumerate several social cues essential for believable agents, including: reflexivity, grouping, attachment, reciprocity, and the ability to attribute mental (intentional) states to oneself and to others.

In~\cite{warpefelt2013analyzing}, the concept of a \emph{Game Agent Matrix} is introduced. This matrix includes columns labeled \emph{single agent}, \emph{multiple agents}, \emph{social structural}, \emph{social goals}, and \emph{cultural historical}, and rows labeled \emph{act}, \emph{react}, and \emph{interact}. The matrix cells incorporate concepts such as \emph{awareness}. The study evaluated many games, finding that numerous concepts, particularly those related to social structural and social goals, were underrepresented. The authors emphasized: 
\begin{quote} ``\textit{NPCs need to exhibit behavior consistent with the environment, 
the situation and their character in order to seem believable.}''
\end{quote}

Incorporating social norms into video games presents substantial challenges. A notable example is \emph{The Elder Scrolls IV: Oblivion}, developed by \emph{Bethesda Software}. In this game, despite the social norm against stealing, players can exploit a loophole by placing a basket over a shopkeeper's head, preventing them from noticing thefts. This is an interesting attempt at introducing realistic perception mechanics, yet it lacks a robust implementation of social norms.\\

\noindent \textbf{Specific Human-Like Activities in the Game.} Let us conclude this section with the observation that games (or virtual environments in general) are highly diverse and open-ended. They may include any aspect of real-world activities, such as conversing in natural language or driving. All such in-game activities can be assessed for their human-likeness~\cite{hecker2020learning}. On one hand, this diversity makes the task of creating general solutions for testing AI believability very challenging. An AI-based \emph{Turing Test} judge should either be customized for a specific activity or capable of evaluating every task humans perform. On the other hand, these specific in-game activities could make it easier for human judges to distinguish between humans and bots. Even if bots are competent and believable in core gameplay, they might fail at some specific tasks, such as engaging in natural conversation or driving on streets.

\section{Experiment Environment}
\label{sec:environment}

\begin{figure}[htb] \centering \includegraphics[height=2.4in]{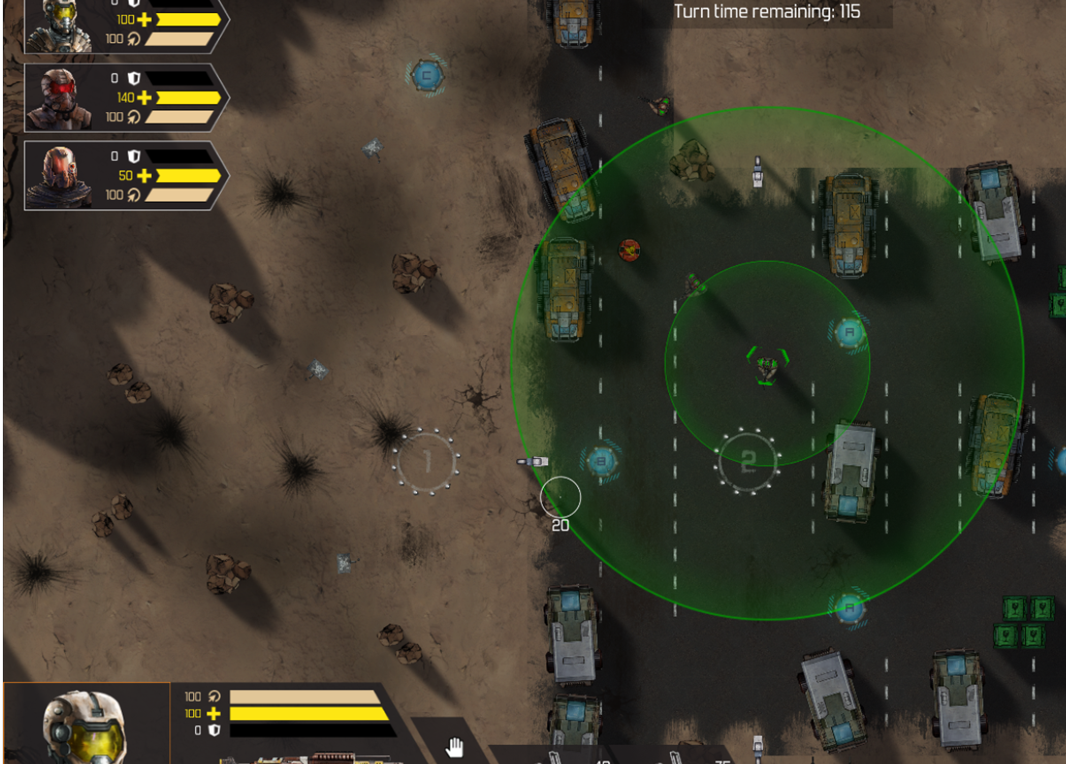} \caption{Movement range in Tactical Troops: Anthracite Shift.} 
\label{fig:tt-movement} 
\Description{
The image shows a top-down view of a Tactical Troops video game map. In the top-left corner, there are portraits representing the player's available units, each displaying three statistics as progress bars: current armor, health, and action points. In the top-right corner, a timer indicates the remaining time for the player's turn. The map itself has two distinct terrain types: the left side consists of dirt ground, while the right side features a large asphalt parking lot with several vehicles.
Three units belonging to the player stand on the parking lot, with one of them currently selected. Around the selected unit, two circular indicators are visible. The smaller circle represents the area where the unit can move while still being able to fire its currently selected weapon, while the larger circle shows the unit’s maximum movement range.
Scattered across the map are various interactive and environmental elements. There are two small areas marked as control points, labeled with the numbers ``1'' and ``2''. Additionally, two pairs of teleporters are present, allowing for rapid movement across different parts of the map. Exploding barrels can also be seen in the bottom-right portion of the map.
}
\end{figure}

The rest of this article is dedicated to a study concerning the problem of believable bots -- their distinction from human players and evaluation of their human-likeness ratio  in a team-based tactical commandos-style game. The game chosen for this study is \emph{Tactical Troops: Anthracite Shift}~\cite{ttroops} (depicted in Fig.~\ref{fig:tt-movement}), which is available on the \emph{Steam} platform\footnote{https://store.steampowered.com/}.


In \emph{Tactical Troops: Anthracite Shift}, each game is played on a 2D map viewed from above. Two players alternate turns, each commanding up to four units. Units are eliminated when they lose all their health points (HP). Unlike many turn-based games, movement is continuous rather than grid-based. The maximum distance a unit can move is determined by its action points (AP), which are also used for performing actions. For example, the larger circle in Fig.~\ref{fig:tt-movement} illustrates the maximum movement range in a single turn, whereas the smaller (inner) circle indicates the range within which a unit can move and still fire its current weapon.

Units can perform actions in any sequence. Considering the main focus of this article, which is the assessment of human-likeness, it is essential to discuss the potential actions each unit can undertake: 
\begin{enumerate} 
\item \textbf{Movement}: Effective movement is crucial as the map's dynamic environments feature buildings for cover, strategic hiding spots, explosive elements, and teleporters that allow for rapid relocation between a designated pair of positions. 
\item \textbf{Shooting}: Each unit is equipped with two weapons from a selection of over 30 types. The shooting action includes choosing a weapon, selecting a shooting mode (single or burst), and positioning accurately (units shoot in the direction they face). 
\item \textbf{Reloading} weapons.
\item \textbf{Using gadgets}: Units carry up to three gadgets, which can be throwable (e.g., grenades, mines) or togglable (e.g., cloaks, shields, and armors). Using throwable gadgets effectively requires careful consideration of positioning and the force applied. 
\item \textbf{Overwatch}: This stance transforms a unit into a stationary defense that automatically fires at the first enemy unit entering its range during the opponent's turn. \end{enumerate}

The game features two alternative victory conditions: 
\begin{enumerate} \item \textbf{Elimination}: Defeat all enemy units. 
\item \textbf{Domination} or \textbf{Devastation}: In Domination, a player wins by controlling the majority of designated areas (control points) on the map for an entire turn. Usually, a map contains three control points, and controlling at least two is necessary for victory. Devastation involves the destruction of specific stationary objects on the map. While the Elimination condition is always applicable, Domination and Devastation modes are exclusive to specific maps. Introducing these modes aims to discourage defensive play (``camping'') and promote more dynamic interactions. 
\end{enumerate}

\subsection{Method Behind AI Agents} For a comprehensive explanation of the AI player's methodology in \emph{Tactical Troops: Anthracite Shift}, please refer to~\cite{ttroops}. This method involves a hybrid approach combining Utility AI~\cite{rabin2013game} and Monte Carlo Tree Search (MCTS)~\cite{mcts}, which are integrated through a blackboard architecture. The Utility AI component handles the strategic layer at a high level. It evaluates and assigns one of six possible orders to each controlled unit, where some orders may include parameters like assaulting or defending a specific point of interest. The utility value of each order is determined by a real-valued numerical score, derived from 10 different considerations (factors). Orders are assigned to units on a greedy basis, meaning that the first unit to receive an order is the one with the globally maximum among each unit's highest scored orders. Assigning an order triggers a recalculation of scores for all other units.

The MCTS algorithm acts as the tactical layer, employing a simplified model of the game. The execution quality of a strategic order is used as a heuristic evaluation function in MCTS, which allows for early termination of simulations. To manage the complexity and avoid combinatorial explosion, the sequence in which units are simulated by MCTS is heuristically determined at each turn. MCTS is allocated a budget of approximately $30,000$ iterations per turn.

\section{Method Behind Human-Likeness Evaluation}
\label{sec:method}

In an initial study~\cite{ttroops}, an XGBoost model utilizing 20 summary features per match was trained to differentiate human players from bots. It achieved an accuracy of $0.68$ and an F1-score of $0.58$. This model was developed using 800 matches and evaluated on an independent set of 200 matches.

\begin{figure}[!ht]
\centering
\includegraphics[height=7.5in]{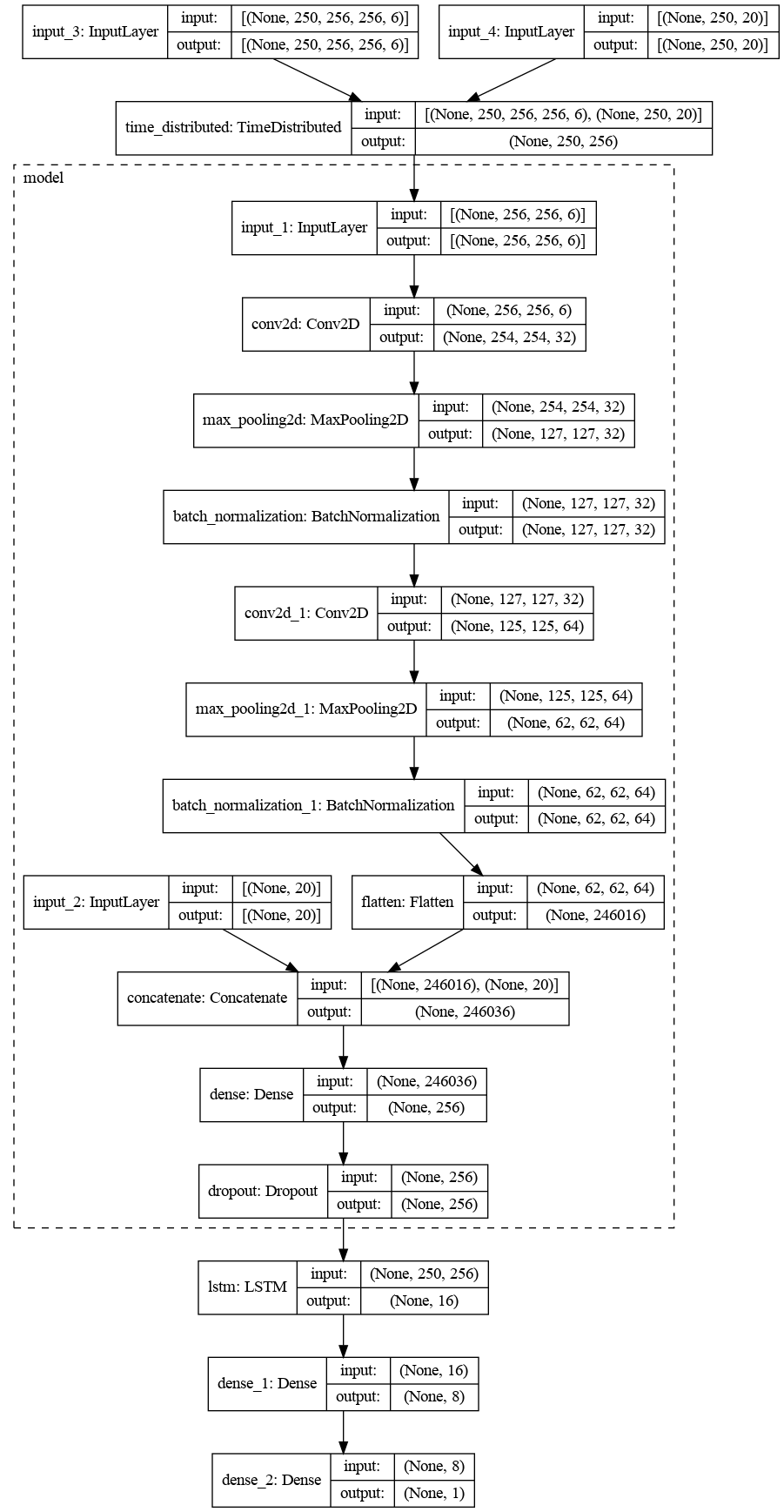}
\caption{Neural Network architecture for the problem.}
\label{fig:nn}
\Description{
The picture represents a diagram of a neural network machine learning model.
The diagram consists of boxes and arrows that illustrate the data flow.
It will be described from the top (input layer) to the bottom (output layer).

The input layer contains two boxes:
(A) Spatial data with dimensions [250, 256, 256, 6], where the last index denotes six different maps.
(B) Flattened scalar features of size [250, 20] for the last 250 states.

The input is connected to a TimeDistributed layer (also referred to as the temporal loop), which repeats the following pattern:
SpatialInput -> Conv2D -> MaxPooling2D -> BatchNormalization -> Conv2D -> MaxPooling2D -> BatchNormalization -> Flatten -> Concatenate (with the Flattened Scalar Features input) -> Dense Layer -> Dropout.

The output from the TimeDistributed layer is subsequently connected to the following components in sequence: 
LSTM -> Dense Layer -> Dense Layer -> Output.

The output represents the probability that the first player encoded in the scalar features in the input is human.
}
\end{figure}

In this study, we present a more sophisticated approach using a hybrid deep neural network that combines convolutional (CNN) and recurrent (RNN) subnetworks. The purpose of the CNN component is to extract features from 2D images, while the RNN component is dedicated to processing summarized data and detecting dependencies within it. The full architecture of the solution is presented in Fig.~\ref{fig:nn}. 

\newpage

We will now give an overview of the input. 

\begin{itemize}
    \item \emph{Input\_1}: graphical (pixel) representation of the map. It involves 6 different layers (submaps): obstacles (see~Fig.~\ref{fig:cnn-input}), rooftops, teleports (that influence movement capabilities), control points, friendly units with health (the health values are normalized and represented as a color saturation), enemy units with health values. The last two maps are generated dynamically, based on the current state of the game (in the current time). 

    \begin{figure}[htb]
    \centering
    \includegraphics[height=2.3in]{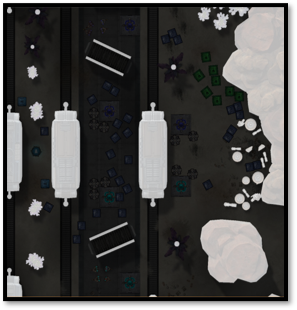}
    \caption{A map of obstacles. One of the six input layers to the spatial component of the neural network. The map of obstacles is crucial because the shapes and positions of obstacles affect lines of sight and lines of fire for the units. Obstacles play a key role in unit placement such as finding hiding spots and avoiding exposure or finding good places to hunt other units. Therefore, they are essential from the perspective of making intelligent positioning decisions.}
    \label{fig:cnn-input}
    \Description{
    A top-down view of the entire level, rendered in a special way to represent a map of obstacles - an idea similar to a mask. The original pixel colors are replaced with either very dark (near-black) or very light (near-white) shades. Light-colored pixels indicate obstacles, representing areas that cannot be moved into or shot through. In this specific map, three relatively large obstacles (train wagons) are positioned in the middle. On the right, two large rock formations serve as additional barriers. Throughout the level, there are also numerous smaller obstacles, such as trees and walls. The background contrasts with these obstacles with dark shades.
    }
    \end{figure}

    \item \emph{Input\_2}: vector representation of 10 x 2 (for both players) numerical features that compute certain values until the current time $t$. These values are: turn number, damage dealt and received, friendly-fire damage, friendly-fire to total damage ratio, \# of used grenades, damage dealt using grenades to total damage, \# of used gadgets, \# of units' status changes.

    \item \emph{Input\_3 and Input\_4}: recurrent neural layers for the spatial and numerical representations, respectively. To properly capture the dynamics of game-play, the input data is passed to the network repeatedly as a sequence of the last 250 states of the game. If a particular game is shorter, then the missing inputs are replaced by zeroes.
\end{itemize}

\subsection{Results}
\label{sec:results}

To evaluate the efficacy of the model, we used 5-fold cross validation on the
set of available game logs. The results are presented in Table~\ref{tab}. 

Let the possible classes be $\left\{ Human, Bot \right \}$. Let $TP_{Human}$ denote the number of true positives calculated for the human class. Analogously, $TN_{X}$, $TN_{X}$, $TN_{X}$ denote true negatives, false positives, and false negatives.
In addition to Macro-F1 score defined as:
\begin{equation}
Macro|F1 = \frac{F1_{Human} + F1_{Bot}}{2}    
\end{equation}
where:
\begin{equation}
    F1_{X} = \frac{2*TP_{X}}{2*TP_{X} + FP_{X} + FN_{X}}
\end{equation}
we also calculated precision for the human class:
\begin{equation}
Precision (Human) = \frac{TP_{Human}}{TP_{Human} + FP_{Human}}
\end{equation}
and recall for the human class:
\begin{equation}
Recall (Human) = \frac{TP_{Human}}{TP_{Human} + FN_{Human}}
\end{equation}
The training data for the game were imbalanced due to the limited availability of human players. It comprised a total of 93,195 logs from 89,667 AI vs. AI matches, 2,190 AI vs. Human matches, and 1,338 Human vs. Human matches. Due to this imbalance, per-class metrics for the human class are included, as this is a minority class and it is more susceptible to higher errors. 

\begin{table}[!ht]
\caption{Results achieved using 5-fold cross validation.}
\label{tab}
\centering
\begin{tabular}{|c|c||c|c|}
\hline
\textbf{Model} & \textbf{F1 Score} & \textbf{Precision (H)} & \textbf{Recall (H)}  \\
\hline\hline
Full (RNN+CNN)& $0.92$ & $0.87$ & $0.81$  \\ \hline
RNN only& $0.88$ & $0.75$ & $0.79$ \\
CNN only& $0.59$ & $0.61$ & $0.57$ \\
\hline
\end{tabular}
\end{table}

The proposed method, that combines CNN and RNN architectures, achieved an F1-Score equal to $0.92$.
This result clearly indicates that the deep learning model significantly outperformed the initial XGBoost model, which was based on only 20 input features, by improving the F1-Score from $0.58$ to $0.92$. 

In Table~\ref{tab}, there is also a comparison of the full approach with models based on a single component only: either the RNN network transforming aggregated scalar features from the game states or the CNN network with six types of 2D maps as input per game state. We can observe that although a considerable amount of information can be inferred from the maps (e.g., the damage dealt thanks to unit health heatmaps or the presence of units at a particular game state), the CNN model performs significantly worse for the task, achieving an F1-score of only $0.59$. The RNN network alone is slightly inferior to the full approach in terms of F1-score but exhibits significantly lower precision. Combining the RNN and CNN components drastically enhances precision and slightly improves the F1-score as well.

Our method achieves better results in differentiating human players from bots compared to most results reported in the literature for different games. For instance, in Pac-Man~\cite{Miranda2017PacManOP}, the differentiation had a success rate of $74\%$. However, it is inappropriate to compare human-like assessment models for agents across different games. As discussed in Section~\ref{sec:challenges}, not only do different environments allow for different expressions of believability, but also the perception of it depends on how the AI was created. For example, both extremely weak and extremely strong AI agents would likely be considered not human-like. The conclusions stemming from these observations will be discussed in the next section.

\balance

\section{Conclusions}
\label{sec:conclusions}

Games have transcended mere entertainment. They also serve as experimental platforms for various algorithms and AI techniques. Developing models to distinguish accurately between humans and AI bots presents unique opportunities in several domains including fake information detection, identity verification (e.g., in educational, healthcare, financial transactions, creative industries, job recruitment, and online dating), scam prevention, and other illegal or immoral activities, along with measuring AI progress. The inspirations trace back to the \emph{Imitation Game}, potentially paving the way for next-generation \emph{CAPTCHAS}.

In the first part of the paper, we explored various aspects involved in the creation and evaluation of believable AI agents. These include the diversity of human players, the complexity of the action spaces, the challenges of scalability, avoiding superhuman capabilities, introducing idle and non-relevant actions, considering biological constraints, incorporating emotional elements, handling uncertainty, adaptability, allowing bots to make and recover from mistakes, training human-like AI, simulating social norms, and challenges related to specific human-like activities within the game.

In the subsequent part, we addressed the challenge of constructing a model for assessing AI in terms of human-likeness within the game \emph{Tactical Troops: Anthracite Shift}. By integrating deep convolutional and recurrent neural networks, we achieved an F1-Score of $0.92$, marking a substantial enhancement from a prior approach that utilized XGBoost and a simpler training configuration. The proposed architecture is relatively general (combining visual spatial data with numeric data), so it can serve as an inspiration for ML models for similar tasks.

We propose an open hypothesis that \textbf{the complexity involved in creating human-like agents in a particular environment correlates inversely with the ease of developing methods to distinguish between humans and bots in that environment}. We invite researchers to engage with this hypothesis to validate or refute it.

We contend that a human-likeness assessment model can serve as an automated quality assurance tool in the development of AI players. This model and the agents can be iteratively refined. The process would start with creating weak bots, which the model would not classify as human-like, and progressively aim to deceive the model through enhancements. Once the model erroneously classifies these improved bots as humans (false positive), it would be judicious to retrain the model.

The presence of human-like NPCs in games offers multiple benefits. Primarily, they enhance the gaming experience by balancing realism and playing strength~\cite{soni2008bots}. Additionally, they can reduce the ``cold start'' effect in massively multiplayer online games until a critical mass of human players is achieved. Furthermore, they facilitate more precise testing by simulating human player behaviors. Human-likeness assessment models also have broader applications, such as bot detection in environments where bot usage by players is considered unfair and is strictly prohibited.

Driven by these various motivations, we will soon organize an open machine learning competition hosted at \emph{knowledgepit.ai} platform. Based on data from \emph{Tactical Troops: Anthracite Shift} provided by us, the task will be to create a model that differentiates human players from bots. The method presented in this article will serve as a baseline.

\bibliographystyle{ACM-Reference-Format} 
\bibliography{bibliography}


\end{document}